\documentclass{article}

\usepackage{listings}
\usepackage{xcolor}
\usepackage{colortbl} 
\usepackage{url}

\usepackage{algorithm}
\usepackage{algorithmic}

\usepackage{enumitem}

\usepackage{amsmath}
\usepackage{mathtools}
\usepackage{amsthm}
\usepackage{amssymb}
\usepackage{bm}
\usepackage{mathrsfs}
\usepackage{multirow}
\usepackage{booktabs} 

\theoremstyle{plain}
\newtheorem{theorem}{Theorem}[section]

\theoremstyle{definition}

\newtheorem{assumption}{Assumption}
\theoremstyle{remark}

\usepackage{changepage}
\definecolor{myhighlight}{RGB}{220,240,255}
\usepackage{multicol}
\usepackage{tcolorbox}
\usepackage{thm-restate}
\usepackage{wrapfig}
\usepackage{threeparttable}

\usepackage{hyperref}

\usepackage{bbm}

\usepackage{microtype}
\usepackage{graphicx}
\usepackage{subfigure}
\usepackage{units}
\usepackage[figuresright]{rotating}

\usepackage[accepted]{icml2021}

\usepackage[capitalize,noabbrev]{cleveref}

\definecolor{codegreen}{rgb}{0,0.6,0}
\definecolor{codegray}{rgb}{0.5,0.5,0.5}
\definecolor{codepurple}{rgb}{0.58,0,0.82}
\definecolor{backcolour}{rgb}{0.95,0.95,0.92}

\tcbset{
  takeaway/.style={
    colback=gray!10!white,
    colframe=gray!60!black,
    fonttitle=\bfseries,
    coltitle=white,
    boxrule=1pt,
    arc=4pt,
    left=6pt,
    right=6pt,
    top=6pt,
    bottom=6pt,
    title=Takeaway,
  }
}
\tcbset{
  example/.style={
    colback=blue!10!white,  
    colframe=blue!60!black, 
    fonttitle=\bfseries,
    coltitle=white,
    boxrule=1pt,
    arc=4pt,
    left=6pt,
    right=6pt,
    top=6pt,
    bottom=6pt,
    title=Example,
  }
}

\begin{document}

\twocolumn[
\icmltitle{PERO: Efficient Robust Post-Training Foundation Models\\ for Encrypted Traffic Classification}
\icmltitlerunning{PERO}
\icmlsetsymbol{correspondence}{*}
\begin{icmlauthorlist}
\icmlauthor{Wumei Du}{nudt}
\icmlauthor{Jiarong Wen}{nudt}
\icmlauthor{Kaiyu Zhang}{nudt}
\icmlauthor{Zi Yang}{nudt}
\icmlauthor{Yiqin Lv}{nudt}
\icmlauthor{Longfei Zhang}{nudt}
\icmlauthor{Dong Liang}{correspondence,nudt}
\icmlauthor{Zheng Xie}{correspondence,nudt}
\end{icmlauthorlist}

\icmlaffiliation{nudt}{National University of Defense Technology, Changsha, China}
\icmlcorrespondingauthor{Dong Liang and Zheng Xie}{dongliangnudt@nudt.edu.cn;xiezheng81@nudt.edu.cn}
\icmlkeywords{Encrypted Traffic Classification, Tail Risk, Supervised Finetuning, Robust Optimization}
\vskip 0.3in
]

\printAffiliationsAndNotice{}

\begin{abstract}
Encrypted traffic classification is vital for network security, yet real-world deployments are inherently sensitive to rare but high-loss errors such as misclassification of malicious traffic.
The encrypted traffic foundation model, as a promising general-purpose technique, can achieve impressive overall performance.
However, employing standard objectives such as empirical risk minimization often overlooks high-risk tail events, and commonly used performance metrics hardly reflect robustness limitations in risk-sensitive scenarios.
Directly applying robust optimization objectives, such as conditional value-at-risk, to post-training is computationally prohibitive for large models, as identifying high-loss samples exhausts substantial computation.
To this end, we propose \textbf{P}re-\textbf{E}valuation \textbf{R}obust \textbf{O}ptimization (\textbf{PERO}), an efficient robust post-training framework for encrypted traffic foundation models.
PERO employs a lightweight proxy to estimate sample-wise risk and selects a subset of high-risk samples to update the foundation model, decoupling risk estimation from expensive large-model optimization.
Extensive experiments on typical encrypted traffic datasets show that PERO achieves competitive or superior robustness and average performance compared to outstanding robust post-training methods, while significantly reducing computational and memory costs.
\end{abstract}





\begin{figure}[ht]
    \centering
    \includegraphics[width=0.95\columnwidth]{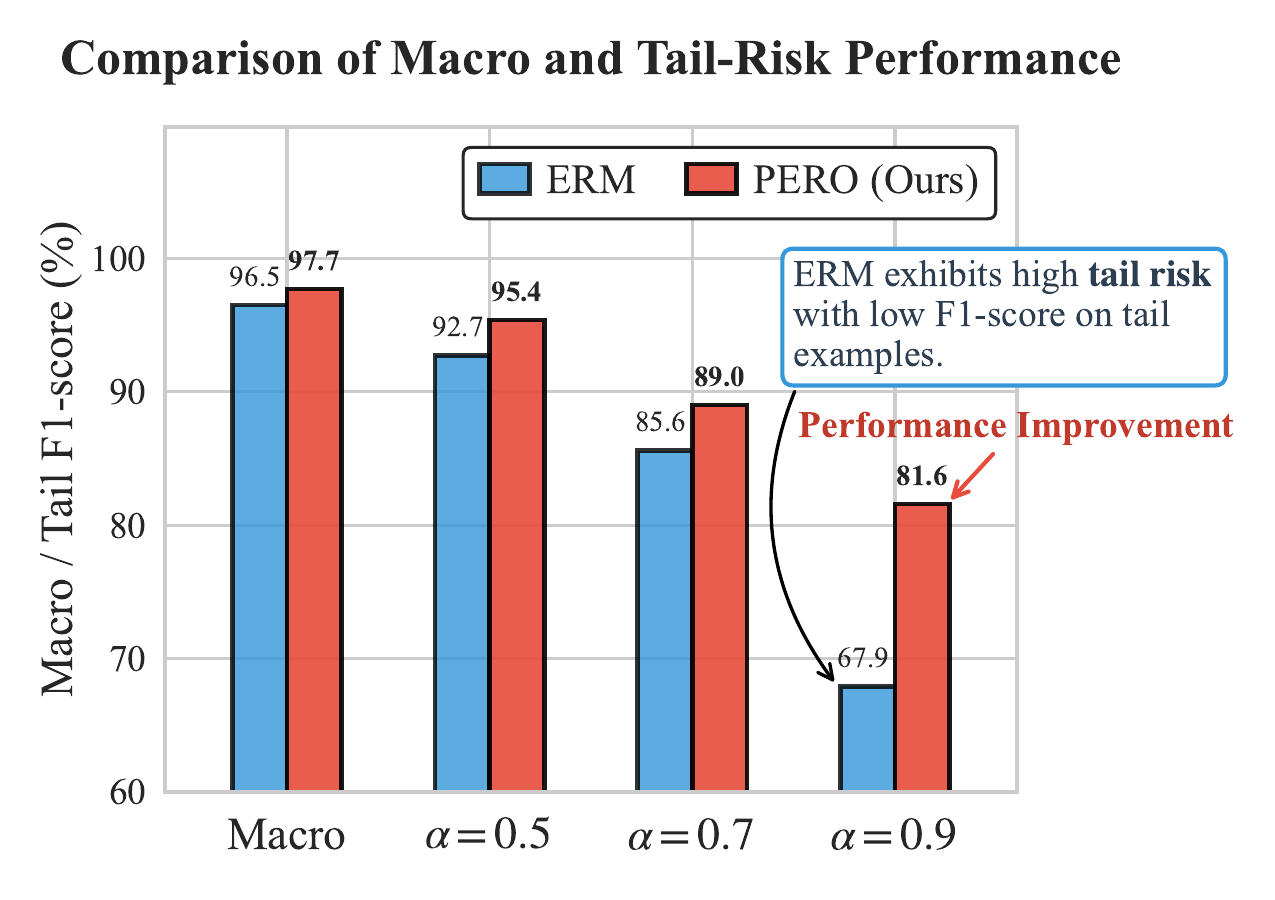}
    \vspace{-0.2cm}
    \caption{Performance comparison between ERM and our method under macro and tail evaluation, with macro denoting class-averaged F1-score and $\alpha$-tail denoting the average F1-score over the worst $(1-\alpha)$ loss-ranked samples.}
    \label{fig:erm_vs_ours}
\end{figure}

\section{Introduction}\label{sec:intro}

In recent years, global Internet usage has surged, driven by advances such as fiber optics, 5G technology, the proliferation of IoT devices, and diverse online services \cite{tahaei2020rise,wang2022machine,azab2024network,Babaria2025fastflow}.
This rapid growth creates both opportunities and challenges for network management, including maintaining quality of service, supporting network planning, and ensuring security \cite{wei2022abl,ARIFFIN2025111471}.
Encrypted traffic classification, which distinguishes normal and abnormal traffic or identifies application types, has thus become increasingly critical \cite{Pacheco2019towards,lotfollahi2020deep,zheng2020learning,zhang2023tfe,wang2025dfe}.
Misclassifying traffic, particularly when it carries malicious payloads, can lead to undetected breaches and cascading service failures.
Consequently, accurate encrypted traffic classification is essential for protecting privacy, maintaining reliable network operations, and enabling timely detection of intrusions and malware \cite{Pacheco2019towards,deldar2023deep,zhang2023automatic}.

Recent advances in encrypted traffic foundation models have yielded substantial performance improvements \cite{zhao2021network,zhao2023yet,dong2025deep,chi2026trustworthy}.
By leveraging self- or weakly supervised pre-training on large-scale unlabeled traffic, models built upon Transformer~\cite{li2025transformer} and BERT-style architectures acquire rich hierarchical representations that capture both packet-level patterns and long-range flow semantics \cite{zhao2022mtflowformer,lin2022bert,meng2023netgpt,qu2024trafficgpt,ZHAN2025110973,shi2026bpfdag}.
These representations can be efficiently adapted to downstream tasks through post-training or fine-tuning with limited labeled data.
Despite these advances, existing post-training procedures predominantly rely on standard empirical risk minimization (ERM) \cite{shai2014understanding}, which prioritizes majority patterns and often overlooks high-risk tail events.
As shown in Fig.~\ref{fig:erm_vs_ours}, this bias inflates tail risk and degrades performance on high-loss samples.
The resulting disparity between average performance and tail robustness underscores the need for robust post-training optimization.

\textbf{Existing Challenges in Robust Post-training Encrypted Traffic Foundation Models:}
Despite the strong empirical performance of encrypted traffic foundation models, achieving robust behavior during post-training remains a fundamental challenge in encrypted traffic classification.
Conditional value-at-risk (CVaR) optimization \cite{rockafellar2000optimization} offers a principled framework for tail-risk minimization by concentrating updates on the fraction of samples with the highest losses, thereby enhancing robustness. 
However, directly integrating such robust optimization into the post-training of foundation models introduces several limitations.

Repeated loss evaluations and parameter updates across millions of parameters impose prohibitive time and energy costs, particularly in scenarios requiring frequent or continual post-training \cite{PEREZJOVE2026111998}. 
Furthermore, memory-limited batch processing restricts exploration of the full loss landscape, resulting in incomplete tail-risk estimation at scale.
Together, these constraints render existing robust optimization frameworks ill-suited for high-throughput network environments, motivating the development of scalable post-training methods that mitigate the misclassification of high-risk encrypted traffic without substantial computational and memory overhead.


\textbf{Developed Method:}
In this paper, we propose \textbf{P}re-\textbf{E}valuation \textbf{R}obust \textbf{O}ptimization (\textbf{PERO}), an efficient robust post-training framework for encrypted traffic foundation models under limited computational and memory constraints.
Rather than designing a new robust optimization objective, we introduce an efficient mechanism within the CVaR framework that decouples risk estimation from full-model parameter updates, enabling tail-focused post-training at significantly reduced computational cost.
Specifically, PERO employs a lightweight pre-evaluation module to estimate sample-wise risk and selects a high-risk subset of samples for updating the foundation model.
The pre-evaluation module is refined online via loss feedback from the foundation model, progressively approximating the underlying risk distribution.
By delegating costly sample screening to a lightweight proxy, PERO preserves the ability to prioritize high-risk samples while substantially reducing computational and memory overhead, making robust post-training of encrypted traffic foundation models more practical.

\textbf{Outline and Primary Contributions:}
The remainder starts with related work in Section \ref{sec:related_work}.
Section \ref{sec:preliminary} introduces notation and preliminaries.
Section \ref{sec:method} presents the proposed method.
Section \ref{sec:exp_analysis} reports experimental results, followed by conclusions.
Our contributions are threefold:
\begin{enumerate}
    \item We propose an efficient robust post-training framework that leverages a lightweight pre-evaluation proxy to guide selective optimization of encrypted traffic foundation models, significantly reducing computational overhead.
    \item We provide a theoretical analysis of PERO, characterizing post-training convergence behavior and deriving a generalization bound for the pre-evaluation module, offering analytical insight into proxy-guided robust post-training.
    \item Extensive experiments on widely used benchmarks demonstrate that our method achieves competitive or superior robustness compared to outstanding robust post-training methods while incurring substantially lower computational and memory costs.
\end{enumerate}

\section{Related Work}\label{sec:related_work}

\subsection{Encrypted Traffic Foundation Models}

Foundation models have become increasingly prominent in encrypted traffic classification, driven by advances in deep learning and the success of Transformer-based architectures such as BERT \cite{dongDeepLearningPretraining2025}.
By first pre-training on massive unlabeled datasets to capture transferable representations and then fine-tuning for specific downstream tasks, these models achieve strong performance across diverse network environments.

Early efforts include PERT \cite{he2020pert}, the first Transformer-based pre-trained model for encrypted traffic, which demonstrated the feasibility of this approach.
ET-BERT \cite{lin2022bert} further improves performance by modeling traffic structure and bidirectional payload associations, reaching over 97\% accuracy across seven tasks.
Building on these foundations, subsequent work addressed architectural and modeling limitations of ET-BERT, leading to specialized designs such as Flow-MAE \cite{hang2023flow}, YaTC \cite{zhao2023yet}, MTC-MAE \cite{xu2024selfsupervised}, TrafficFormer \cite{zhou2025trafficformer}, and CETP \cite{lin2024cetp}.
In parallel, generative pre-training has led to NetGPT \cite{meng2023netgpt}, TrafficGPT \cite{qu2024trafficgpt}, and the encoder-decoder-based Lens \cite{wang2024lens}, while NetMamba \cite{wang2024netmamba} replaced the Transformer with an optimized Mamba architecture for efficient online classification. 
More recently, MIETT \cite{chen2025miett} introduces a two-level attention transformer to jointly capture intra-packet and inter-packet dependencies, outperforming prior token-centric foundation models on multiple benchmarks.

Despite their strong overall performance, these models are typically optimized with standard empirical objectives that underemphasize rare high-loss samples, while robust optimization during post-training remains prohibitively costly at this scale.

\subsection{Robust Optimization for Large-Scale Models}

Robust optimization has evolved from foundational theory to widely used methods in machine learning, aiming to train models that are reliable not only on average but also under worst-case conditions.
Typical approaches are rooted in the distributionally robust optimization (DRO) framework, which addresses uncertainty in data distributions and was originally formalized with convex uncertainty and moment-based ambiguity sets \cite{ben1998robust,ben2002robust,delage2010distributionally,Wiesemann2014Distributionally,zou2025utility}.
Representative methods include CVaR optimization \cite{rockafellar2000optimization,wang2023simple,lv2024theoretical, wang2026model, qu2025fast,wang2025robust}, GroupDRO \cite{Sagawa2020Distributionally}, and recent LogSumExp-based DRO formulations such as TDRO \cite{gladin2025improved}.
CVaR optimization minimizes the expected loss in the worst-performing tail of the data distribution, providing robustness to high-risk events.
GroupDRO focuses on ensuring uniform accuracy across predefined subpopulations, thereby preventing the model from overfitting to the majority groups while neglecting minority groups.
TDRO uses a smooth LogSumExp surrogate to emphasize high-loss samples and improve tractability in stochastic optimization.
Related hard-sample mining strategies also aim to focus training on difficult examples.
For example, online hard task mining (OHTM) \cite{kumar2023effect} maintains a buffer to store past difficult data points and integrates them into the training pipeline.
By emphasizing worst-case performance, robust optimization offers a principled way to improve reliability when standard ERM underperforms.

In post-training large foundation models, CVaR incurs substantial computational and memory overhead from repeated tail-loss estimation and re-optimization, while memory-limited batching further degrades tail-risk estimation at scale.
GroupDRO relies on predefined and often static group structures that may be misaligned with evolving or latent risk patterns, limiting its robustness in large-scale settings.
TDRO's scalability to budget-limited foundation-model post-training remains unexplored.
OHTM heuristically selects high-loss samples without explicit tail-risk modeling and exhibits unstable behavior under nonstationarity.

\section{Preliminary}\label{sec:preliminary}
\subsection{Notations}\label{subsec:notations}
Let $\mathbf{x}\in\mathcal{X}$ denote an input sample used for classification, and let $y\in\mathcal{Y}$ be its corresponding label.
The classification model is parameterized by $\bm\theta\in\bm\Theta$, while the underlying data distribution is denoted by $p(\mathbf{x},y)$.  
We consider a dataset $\mathcal{D}=\{(\mathbf{x}^{(i)},y^{(i)})\}_{i=1}^{N}$ with $N$ training sample pairs.
The objective of encrypted traffic classification is to learn a mapping $f_{\bm\theta}:\mathcal{X}\to\mathcal{Y}$ that minimizes classification errors, typically expressed as $\mathbb{E}_{p(\mathbf{x},y)}\big[\mathbbm{1}_{\{f_{\bm\theta}(\mathbf{x})\neq y\}}\big]$, where $\mathbbm{1}_{\{\cdot\}}$ is the indicator function.
For optimization, we adopt the cross-entropy loss, denoted by $\ell(\cdot,\cdot)$, and write
\[
\ell=\ell(f_{\bm\theta}(\mathbf{x}), y),\quad \ell^{(i)}=\ell(f_{\bm\theta}(\mathbf{x}^{(i)}), y^{(i)}),
\]
where $\ell$ is the loss for a generic sample pair $(\mathbf{x}, y)$, and $\ell^{(i)}$ is the loss for the $i$-th training sample pair in $\mathcal{D}$.

\subsection{ERM and Robust Optimization Methods}

\paragraph{Empirical Risk Minimization (ERM)}
ERM is the standard principle for supervised learning, aiming to find a predictor that minimizes expected loss under the true data distribution $p(\mathbf{x},y)$ \cite{shai2014understanding}.
Since $p(\mathbf{x},y)$ is unknown, ERM minimizes the loss under the empirical distribution $\hat{p}$ induced by the training data:
\begin{equation}
\begin{split}\label{eq:erm}
\min_{\bm\theta\in\bm\Theta} \mathbb{E}_{\hat{p}(\mathbf{x},y)} \bigl[ \ell \bigr]=\min_{\bm\theta\in\bm\Theta} \frac{1}{N} \sum_{i=1}^N \ell^{(i)}.
\end{split}
\end{equation}

\paragraph{Conditional Value-at-Risk (CVaR)}
CVaR focuses on the upper tail of the loss distribution.
For a risk level $\alpha \in (0,1)$, e.g., $\alpha=0.9$, corresponding to the worst 10\% of cases, it minimizes the expected loss among the worst-performing $(1-\alpha)$ fraction of samples \cite{rockafellar2000optimization,rockafellar2002conditional}:
\begin{equation}
\begin{split}
\min_{\eta\in\mathbb{R}}\;
    \eta + \frac{1}{1-\alpha} \mathbb{E}\big[(\ell-\eta)_{+}\big],
\end{split}
\end{equation}
where $(a)_{+}=\max(a,0)$, and $\eta$ approximates the Value-at-Risk(VaR) threshold.
For large-scale applications, CVaR is commonly optimized using \textbf{Monte Carlo-based stochastic approximation} \cite{tamar2015optimizing}, which we refer to as \textbf{MC-CVaR} for brevity.
Specifically, given a mini-batch $\{(\mathbf{x}^{(i)},y^{(i)})\}_{i=1}^{B}$ with losses $\ell^{(i)}$, the stochastic subgradient of $\mathrm{CVaR}_{\alpha}$ with respect to $\bm\theta$ is estimated as:
\begin{equation}
\begin{split}
\nabla_{\bm\theta}\text{CVaR}_\alpha \approx\frac{1}{(1-\alpha)B}\sum_{i=1}^{B}\mathbbm{1}_{\{\ell^{(i)}>\eta\}}\nabla_{\bm\theta}\ell^{(i)},
\end{split}
\end{equation}
with the threshold $\eta$ iteratively updated to approximate the $\alpha$-quantile of the loss distribution.

\paragraph{Group Distributionally Robust Optimization (GroupDRO)}
GroupDRO aims to learn model parameters $\bm\theta$ that minimize the maximum group-wise expected loss \cite{Sagawa2020Distributionally}.
Given a dataset split into $G$ groups $\mathcal{D} = \bigcup_{g=1}^{G} \mathcal{D}_g$, where each group $g$ indexes a subpopulation, e.g., application types, GroupDRO solves the min-max problem:
\begin{equation}
\begin{split}
\min_{\bm\theta}\;\max_{g\in[G]} \mathbb{E}_{(x,y)\sim\mathcal{D}_g}\bigl[\ell\bigr],
\end{split}
\end{equation}
where the inner maximum seeks the group with the highest expected loss, focusing on the worst-performing subpopulation.

\subsection{From ERM to Robust Post-training}
Given a large-scale pre-trained foundation model $f_{\bm\theta}$ and a downstream encrypted traffic classification task defined by the data distribution $p_{\text{task}}(\mathbf{x},y)$, the post-training stage adapts the model parameters $\bm\theta$ to the target task using a labeled dataset
\[
\mathcal{D}_{\text{task}} = \{(\mathbf{x}^{(i)},y^{(i)})\}_{i=1}^{N},\quad (\mathbf{x}^{(i)},y^{(i)})\sim p_{\text{task}}(\mathbf{x},y).
\]
The standard post-training objective follows ERM, as shown in Eq.~\eqref{eq:erm}, and is typically optimized via stochastic gradient descent.
ERM effectively adapts pre-trained representations when the training data $\mathcal{D}_{\text{task}}$ is representative of the target distribution.
However, this assumption is often violated in encrypted traffic classification, where high-risk traffic examples are significantly underrepresented.
As a result, ERM disproportionately favors majority patterns, leading to severe performance degradation on high-risk tail examples.
These limitations motivate robust post-training objectives that explicitly account for tail risk, aiming to ensure reliable classification performance even in high-risk traffic scenarios.

\section{Method}\label{sec:method}

Robust post-training of encrypted traffic foundation models incurs substantial computational overhead.
Identifying high-risk samples typically requires full-model inference on large candidate batches, a process rendered prohibitively expensive by the model's scale and complexity.
To address this challenge, we propose \textbf{P}re-\textbf{E}valuation \textbf{R}obust \textbf{O}ptimization (\textbf{PERO}), a resource-efficient framework for scalable robust post-training.

\subsection{Overall Framework of PERO}

\begin{figure*}[ht]
\begin{center}
\centerline{\includegraphics[width=0.99\textwidth]{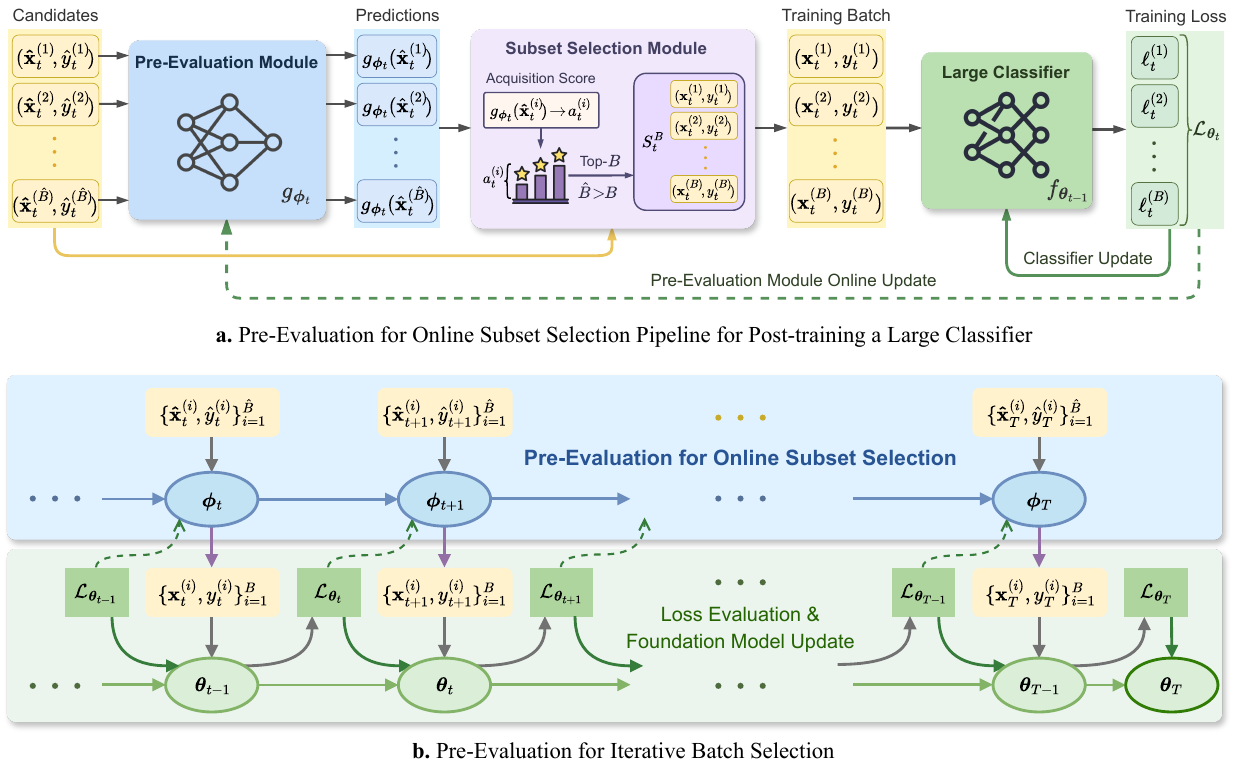}}
\caption{Overview of PERO Framework.
Panel (a) shows the optimization loop with the pre-evaluation module $\bm\phi$, subset selection module, and classifier $\bm\theta$, with $|\bm\phi| \ll |\bm\theta|$.
Panel (b) depicts iterative batch selection guided by the pre-evaluation module.
}
\label{pipeline}
\end{center}
\vspace{-4pt}
\end{figure*}

PERO approximates robust post-training by using a lightweight proxy to guide selective optimization of the foundation model.
We study encrypted traffic classification as a representative application because it is risk-sensitive, and robust post-training of large models incurs substantial computational overhead.
The framework consists of a pre-evaluation module for risk estimation and a subset selection module that identifies risk-informative samples for updating the large classifier.
The overall learning pipeline proceeds as follows (see Fig.~\ref{pipeline}):
(i) the pre-evaluation module is updated using sample-wise losses from the previous iteration to approximate the classifier's current loss landscape;
(ii) the module predicts losses for a candidate pool of size $\hat{B}$ and identifies samples likely to contribute to tail risk;
(iii) the subset selection module selects the $B$ samples ($B<\hat{B}$) with the highest predicted risk scores from the pool at each iteration to update the classifier.

\subsection{Pre-Evaluation Module Formulation}
The pre-evaluation module works as an online surrogate risk estimator that predicts sample-wise classification risk under the current model state.
The module takes in the sample and aims to predict the cross-entropy loss induced by the evolving classifier, approximating the loss mapping $f: \bm\Theta\times\mathcal{X}\to\mathbb{R}_{+}$, which evolves as the classifier parameters $\bm\theta$ are updated iteratively.
As cross-entropy reflects both prediction error and classifier confidence, the predicted loss serves as a principled proxy for the instantaneous risk and reveals sample-level uncertainty.
This formulation casts pre-evaluation as a supervised learning problem that is intrinsically entangled with the main classifier's training dynamics.
Consequently, loss prediction is non-trivial, as the continual updates to $\bm\theta$ create a non-stationary and evolving risk landscape.

We formulate the pre-evaluation module as a lightweight neural network $g : \bm\Phi \times \mathcal{X} \rightarrow \mathbb{R}_{+}$, parameterized by $\bm\phi \in \bm\Phi$ and updated iteratively.
Let $(\mathbf{\hat x}_t^{(i)}, {\hat y}_t^{(i)}) \in \mathcal{X} \times \mathcal{Y}$ denote the $i$-th sample pair of the batch chosen at iteration $t$.
The module $g_{\bm\phi_{t}}$ takes a sample $\mathbf{\hat x}_t^{(i)} \in \mathcal{X}$ as input and outputs an estimated loss value $\hat{g}_t^{(i)}$.
Formally,
\[ 
g : \bm\Phi \times \mathcal{X} \rightarrow \mathbb{R}, \quad (\bm\phi_{t}, \mathbf{\hat x}_t^{(i)}) \mapsto \hat{g}_t^{(i)} \coloneqq g_{\bm\phi_{t}}(\mathbf{\hat x}_t^{(i)}),
\]
where $\hat{g}_t^{(i)}$ serves as an estimate of the true cross-entropy loss ${\hat \ell}_t^{(i)}\coloneqq\ell(f_{\bm\theta_{t-1}}(\mathbf{\hat x}_t^{(i)}), {\hat y}_t^{(i)})$ that $(\mathbf{\hat x}_t^{(i)}, {\hat y}_t^{(i)})$ would incur under the classifier parameters $\bm\theta_{t-1}$.
This mechanism facilitates exploration by highlighting samples with higher prediction risk, often located in regions of uncertainty or tail risk, thereby guiding the classifier to improve robustness under worst-case scenarios.

The pre-evaluation module is trained in a supervised manner to approximate the sample-wise classification loss.
At iteration $t$, the classifier $f_{\bm\theta_{t-1}}$ yields the true losses for a batch $\{(\mathbf{x}_t^{(i)},y_t^{(i)})\}_{i=1}^{B}$:
\[
\ell_t^{(i)}\coloneqq\ell(f_{\bm\theta_{t-1}}(\mathbf{x}_t^{(i)}), y_t^{(i)}),\quad i=1,\cdots,B.
\]
The module parameters $\bm\phi_{t}$ are updated online by minimizing the mean squared error (MSE) between predictions $g_t^{(i)}\coloneqq g_{\bm\phi_{t}}(\mathbf{x}_t^{(i)})$ and true losses $\ell_t^{(i)}$ via gradient descent with a learning rate $\gamma_{\phi}$:
\begin{equation}
\label{eq:pre-mse}
\mathcal{L}_{\text{MSE}}(\bm{\phi}_{t}) = \frac{1}{B} \sum_{i=1}^{B} \big[g_t^{(i)} - \ell_t^{(i)}\big]^2,\ \bm{\phi}_{t+1} = \bm{\phi}_{t} - \gamma_{\phi}\nabla_{\bm{\phi}} \mathcal{L}_{\text{MSE}}(\bm{\phi}_{t}).
\end{equation}
Through this iterative online optimization, the pre-evaluation module progressively learns a mapping from input samples to their expected classification difficulty under the evolving model state.

\subsection{Pre-Evaluation for Online Subset Selection}

To improve robustness while reducing computational cost, the subset selection module selects critical samples online at each iteration to update the large classifier.
Each training batch is expanded into a candidate pool of size $\hat{B} > B$.
The pre-evaluation module predicts loss estimates for all candidates, which are then used to select a subset of $B$ samples with the highest potential to improve classifier performance in uncertain or high-risk regions of the data space.

Formally, let $\{\hat{\mathbf{x}}_{t}^{(i)}\}_{i=1}^{\hat B}$ denote the candidate samples at iteration $t$.
The pre-evaluation module outputs a prediction $\hat{g}_t^{(i)}=g_{\bm\phi_{t}}(\mathbf{\hat x}_t^{(i)})$, which is then assigned an acquisition score $a_t^{(i)}$ that quantifies its informativeness or difficulty relative to the current classifier $f_{\bm\theta_{t-1}}$.
We introduce a binary indicator vector $\mathbf{z}_t = [z_t^{(1)}, \dots, z_t^{(\hat{B})}]^\top \in \{0, 1\}^{\hat{B}}$, where $z_t^{(i)}=1$ indicates that the candidate $\hat{\mathbf{x}}_t^{(i)}$ is selected, and $z_t^{(i)}=0$ indicates that it is not.
From a general perspective, the subset selection at iteration $t$ can be formulated as a constrained set-function maximization problem:
\begin{equation}
\setlength{\jot}{-6pt}
\begin{aligned}
\max_{\mathbf{z}_t \in \{0, 1\}^{\hat{B}}} \quad & \mathcal{A}(\mathbf{z}_t) = \underbrace{\sum_{i=1}^{\hat{B}} z_t^{(i)} \cdot a_t^{(i)}}_{\text{Utility}} + \lambda \cdot \underbrace{\mathcal{S}(\{z_t^{(i)}\mathbf{\hat x}_t^{(i)}\}_{i=1}^{\hat{B}})}_{\text{Set-level Bonus}} \\
\text{s.t.} \quad & \sum_{i=1}^{\hat{B}} z_t^{(i)} \cdot c_t^{(i)} \leq C, \quad z_t^{(i)} \in \{0, 1\}.
\end{aligned}
\end{equation}
Here, $\mathcal{S}(\{z_t^{(i)}\mathbf{\hat x}_t^{(i)}\}_{i=1}^{\hat{B}})$ captures additional set-level incentives such as the coverage measure of $\mathcal{X}$ space, $\lambda$ controls the trade-off between utility and bonus, $c_t^{(i)}$ denotes the cost of the $i$-th sample used for modeling resource constraints, and $C$ represents the total resource budget per optimization iteration during post-training.

The subset selection strategy in PERO is an efficient instantiation of the general formulation.
Since all preprocessed samples have uniform processing cost, we set $c_t^{(i)}=1$ and enforce $\sum_{i=1}^{\hat{B}} z_t^{(i)}=B$.
To avoid solving a complicated robust optimization problem, we set $\lambda=0$, reducing the objective to top-$B$ selection by acquisition score.
Meanwhile, PERO uses the predicted loss $\hat{g}_t^{(i)}$ from the pre-evaluation module as the acquisition score $a_t^{(i)}$.
Since cross-entropy loss reflects predictive uncertainty and worst-case risk, higher predicted losses correspond to samples that are more likely to lie in uncertain or underexplored regions of the input space.
By prioritizing these examples, PERO selects $S_t^B = \arg\max_{\mathbf{z}_t} \mathcal{A}(\mathbf{z}_t)$ and focuses post-training on challenging samples while maintaining computational efficiency.

Empirically, this straightforward top-$B$ criterion yields substantial robustness gains and consistently outperforms alternative robust optimization methods with minimal overhead.
Our results show that effective robust optimization can be achieved without resorting to complex subset selection objectives.
The framework is also flexible, allowing more sophisticated selection criteria to be explored in future work.

\subsection{Algorithm Overview}
Building on the subset selection mechanism described above, the optimization restricts parameter updates to the selected high-risk subset.
At iteration $t$, we perform a standard stochastic gradient descent step on the selected subset $S_t^B=\{(\mathbf{x}_t^{(i)},y_t^{(i)})\}_{i=1}^B$ to update the classifier parameters from $\bm\theta_{t-1}$ to $\bm\theta_{t}$:
\begin{equation}\label{eq: opt step}
\begin{split}
\bm\theta_{t}=\bm\theta_{t-1} - \frac{\gamma_{\theta}}{B}\sum_{i=1}^{B}\nabla_{\bm\theta} \ell_{t}^{(i)},
\end{split}
\end{equation}
where $\gamma_{\theta}$ is the learning rate.
This update concentrates optimization on high-risk samples while avoiding full-batch updates.
The complete training procedure is summarized in Algorithm \ref{alg}.

\begin{algorithm}[t!]
    \caption{Pre-Evaluation Subset Selection in the Post-Training of a Large Pre-Trained Classifier}
    \begin{algorithmic}[1]
        \STATE \textbf{Input}:
        Task dataset $\mathcal{D}_{\text{task}}$;
        pre-trained classifier $f_{\bm\theta_0}$;\\
        iterations $T$;
        batch size $B$;
        candidate batch size $\hat{B}$;
        learning rates $\gamma_{\phi}$, $\gamma_{\theta}$.
        \STATE \textbf{Output}:
        Optimized classifier $f_{\bm\theta_T}$.
        \STATE Initialize pre-evaluation module $g_{\bm\phi_0}$.
        \STATE Randomly sample an initial batch $\{(\mathbf{x}_0^{(i)}, y_0^{(i)})\}_{i=1}^{B} \sim \mathcal{D}_{\text{task}}$.
        \STATE Compute initial losses $\{\ell_0^{(i)} \coloneqq \ell(f_{\bm\theta_0}(\mathbf{x}_0^{(i)}), y_0^{(i)})\}_{i=1}^{B}$ and pre-evaluation module predictions $\{g_0^{(i)}\coloneqq g_{\bm\phi_0}(\mathbf{x}_0^{(i)})\}_{i=1}^{B}$.
        \FOR{$t = 1 : T$}
        \STATE{\textcolor{blue}{\textit{// Pre-Evaluation Module Online Update}}}
        \STATE Update $\bm\phi_{t}$ from $\bm\phi_{t-1}$ using $\{\ell_{t-1}^{(i)}, g_{t-1}^{(i)}\}_{i=1}^{B}$ via Eq.\eqref{eq:pre-mse}:\\
        $\displaystyle \bm\phi_{t} \leftarrow \bm\phi_{t-1} - \gamma_{\phi} \nabla_{\bm\phi}\mathcal{L}_{\text{MSE}}(\bm\phi_{t-1})$.
        \STATE{\textcolor{blue}{\textit{// Active Subset Selection}}}
        \STATE Randomly sample a candidate batch $\{(\mathbf{\hat x}_{t}^{(i)}, \hat y_{t}^{(i)})\}_{i=1}^{\hat B} \sim \mathcal{D}_{\text{task}}$.
        \STATE Predict losses: ${\hat g}_{t}^{(i)}=g_{\bm\phi_{t}}(\mathbf{\hat x}_{t}^{(i)}), \forall i \in \{1,\dots,\hat B\}$.
        \STATE Assign acquisition scores: $a_{t}^{(i)} \!\leftarrow {\hat g}_{t}^{(i)}, \forall i \in \{1,\dots,\hat B\}$.
        \STATE Select the top-$B$ samples with the largest $a_{t}^{(i)}$ to form $S_{t}^B=\{(\mathbf{x}_{t}^{(i)},y_{t}^{(i)})\}_{i=1}^{B}$.
        \STATE{\textcolor{blue}{\textit{// Loss Evaluation and Foundation Model Update}}}
        \STATE Compute true losses: $\{\ell_{t}^{(i)}\coloneqq \ell(f_{\bm\theta_{t-1}}(\mathbf{x}_{t}^{(i)}), y_{t}^{(i)})\}_{i=1}^{B}$.
        \STATE Record module predictions $\{g_t^{(i)}\coloneqq g_{\bm\phi_t}(\mathbf{x}_t^{(i)})\}_{i=1}^B$.
        \STATE Update the classifier parameters on $S_{t}^B$ via Eq.\eqref{eq: opt step}:\\
        $\displaystyle \bm\theta_{t} \leftarrow \bm\theta_{t-1} - \frac{\gamma_{\theta}}{B}\sum\nolimits_{i=1}^B\nabla_{\bm\theta} \ell_{t}^{(i)}$.
        \ENDFOR
    \end{algorithmic}
    \label{alg}
\end{algorithm}

\subsection{Theoretical Analysis}

We provide a theoretical analysis of PERO under simplifying assumptions, focusing on
(i) the convergence behavior of the proposed post-training procedure and
(ii) a generalization bound for the pre-evaluation module.
The results provide analytical insight into proxy-guided post-training rather than a full characterization of nonconvex foundation-model optimization.
Full proofs are provided in the online supplement.\footnote{\url{https://anonymous.4open.science/r/PERO-D0A7}}

\textbf{Preliminaries and Assumptions.}
We adopt the notation from Section~\ref{subsec:notations} and assume access to $n$ i.i.d. samples $\{(\mathbf{x}_{i},y_{i})\}_{i=1}^{n}$ drawn from the distribution $p(\mathbf{x},y)$.
To analyze the convergence of the coupled updates $\{{\bm \theta}_t, {\bm \phi}_t\}_t$ and the generalization bound of the pre-evaluation module, we impose the following assumptions on the functions ${\ell}(f_{\bm\theta}({\mathbf x}), y) $ and $g_{\bm\phi}({\mathbf x})$.

\begin{assumption}[Bounded Loss]
\label{assump:bounded}
$\ell(f_{\bm\theta}(\mathbf{x}),y)$ is uniformly bounded: $\ell(f_{\bm\theta}(\mathbf{x}),y) \le L_{\max}$ for all $(\mathbf{x},y)$.
\end{assumption}

\begin{assumption}[Bounded Spectral Radius]
\label{assump:bounded_SR}
$\ell(f_{\bm\theta}(\mathbf{x}),y)$ is twice continuously differentiable with respect to $\bm\theta$, and its Hessian satisfies $\lambda_{\max}\big(\nabla^2_{\bm\theta\bm\theta}\ell(f_{\bm\theta}(\mathbf{x}),y)\big) \le 1$ for all $(\mathbf{x},y)$.
\end{assumption}
 
\begin{assumption}[Lipschitz Continuity]\label{assump:Lipschitz}
$ g_{\bm \phi}({\mathbf x}) $ is Lipschitz continuous with respect to $\bm\phi$, i.e., $| g_{\bm{\phi}_1}(\mathbf{x}) - g_{\bm{\phi}_2}(\mathbf{x}) | \le L \|\bm{\phi}_1 - \bm{\phi}_2\|_2$ for all $\mathbf{x}$ and some constant $L>0$.
\end{assumption}

\textbf{Convergence Analysis.}
Building on the above assumptions, we analyze the convergence of the proposed post-training procedure. Theorem~\ref{thm:convergence} shows that, under mild conditions, the parameter sequence $\{{\bm \theta}_t, {\bm \phi}_t\}_t$ converges to a stationary point.

\begin{theorem}[Convergence to Stationary Points]
\label{thm:convergence}
Under Assumptions~\ref{assump:bounded}, \ref{assump:bounded_SR} and \ref{assump:Lipschitz}, suppose that the prediction error holds: $ g_{\bm \phi}({\mathbf x}^{(i)}) - \ell^{(i)} < \delta, \forall i $ for a constant $\delta>0$, and that the spectral radius of $ \nabla^2_{\bm{\phi\phi}}{g}({\mathbf x}) $ is bounded by $\frac{1}{\delta}\left(\frac{1}{2} - L^2\right)$.
Then, the sequence $ \{{\bm \theta}_t, {\bm \phi}_t\}_t $ converges to a stationary point $({\bm \theta}^*, {\bm \phi}^*)$.
\end{theorem}

We note that Theorem~\ref{thm:convergence} relies on a uniformly bounded loss-Hessian spectral radius throughout optimization.
This is a standard smoothness condition widely used in convergence analyses of gradient-based methods~\cite{hu2023beyond}, although its global validity can be restrictive for large Transformer models.
Accordingly, we use this assumption to obtain a tractable analytical characterization of the coupled update dynamics, and interpret Theorem~\ref{thm:convergence} as an insight under explicit assumptions rather than a complete description of practical nonconvex optimization.

\textbf{Generalization Bound.}
Denote by $ R({\bm \theta}^*, {\bm \phi}^*)$ the expectation $\mathbb{E}_{p({\mathbf x}, y)}\!\!\left[\ell\right] $, and let $ \widehat{R}({\bm \theta}^*, {\bm \phi}^*) = \frac{1}{\hat{B}}\sum_{i=1}^{\hat{B}} z^{(i)} \ell^{(i)} $ and $ \widetilde{R}({\bm \theta}^*) = \frac{1}{\hat{B}}\sum_{i=1}^{\hat{B}} \ell^{(i)} $.
Theorem \ref{thm:GB} addresses the following key question regarding generalization:
How well does the learned risk estimator perform on unseen data?


\begin{theorem}[Generalization Bound for the Pre-Evaluation Module]
\label{thm:GB}
Under Assumption~\ref{assump:bounded}, the following generalization bound holds
with a probability of at least $1-\epsilon$ for any $\epsilon\in(0,1)$:
\begin{align*}
\left|
R({\bm \theta}^*,{\bm \phi}^*)
-
\widehat{R}({\bm \theta}^*,{\bm \phi}^*)
\right|
&<
\sqrt{
\left(
\mathbb{V}[\ell]
+
\frac{L_{\max}}{3}
\right)
\frac{
2\ln\left(\frac{1}{\epsilon}\right)
}{
\hat{B}
}
}
\\
&\quad
+
L_{\max}
\sqrt{
\frac{\hat{B}}{2B^2}
\ln\left(\frac{1}{\epsilon}\right)
}
\\
&\quad
+
\frac{
L_{\max}(\hat{B}-B)
}{
\hat{B}
}.
\end{align*}
with $\mathbb{V}[\cdot]$ denoting variance and $L_{\max}$ as in
Assumption~\ref{assump:bounded}.
\end{theorem}

In conjunction with the confidence $\epsilon$ and a sufficiently large candidate batch $\hat{B}$, Theorem~\ref{thm:GB} reveals the generalization bound given the learned parameters $(\bm\theta^*, \bm\phi^*)$.
It is also associated with the variance $\mathbb{V}_{\tau_i\sim p_\alpha(\tau)}[{\ell}(f_{{\bm \theta}^*}({\mathbf x}), y)]$.
This bound establishes a worst-case generalization guarantee under standard i.i.d. assumptions, and serves as a theoretical upper envelope for the non-stationary iterative process, ensuring the proxy's selection error remains bounded.

\subsection{Computational Complexity Analysis}

We analyze the per-iteration computational complexity of PERO relative to ERM and MC-CVaR.
Let $\mathcal{C}_{f}$ and $\mathcal{C}_{f}^{\text{fwd}}$ denote the forward-backward and inference costs of the foundation model $f_{\bm\theta}$, and similarly $\mathcal{C}_{g}, \mathcal{C}_{g}^{\text{fwd}}$ for the pre-evaluation module $g_{\bm\phi}$, with $\mathcal{C}_{g} \ll \mathcal{C}_{f}$.

Table~\ref{tab:complexity} summarizes the per-iteration complexity.
ERM is most efficient but not robust, whereas MC-CVaR introduces a costly $\mathcal{O}(\hat{B} \cdot \mathcal{C}_{f}^{\text{fwd}})$ large-model screening bottleneck.
PERO mitigates this bottleneck by delegating sample selection to a lightweight proxy.
Since ranking is negligible and $\mathcal{C}_g, \mathcal{C}_g^{\mathrm{fwd}} \ll \mathcal{C}_f$, PERO achieves robust screening with computational overhead close to ERM.

\begin{table}[h]
\centering
\caption{Per-iteration computational complexity.}
\label{tab:complexity}
\scalebox{0.9}{
\begin{tabular}{ll}
\toprule
Method & {Computational Complexity} \\
\midrule
ERM & $\mathcal{O}(B \cdot \mathcal{C}_{f})$ \\
MC-CVaR & $\mathcal{O}(B \cdot \mathcal{C}_{f} + \hat{B} \cdot \mathcal{C}_{f}^{\text{fwd}})$ \\
{PERO (Ours)} & $\mathcal{O}(B \cdot \mathcal{C}_{f} + \hat{B} \cdot \mathcal{C}_{g}^{\text{fwd}} + B \cdot \mathcal{C}_{g})$ \\
\bottomrule
\end{tabular}}
\end{table}

\section{Experimental Results and Analysis}\label{sec:exp_analysis}

In this section, we conduct extensive experiments on four typical encrypted traffic classification datasets (Section~\ref{subsec:setting}) to evaluate the effectiveness, robustness, and efficiency of PERO.
We compare PERO with representative robust post-training and sample-selection baselines (Section~\ref{subsec:comparison}), examine the risk prediction fidelity of the pre-evaluation module (Section~\ref{subsec:predict}), and perform ablation studies to assess the impact of key hyper-parameters (Section~\ref{subsec:ablation}).

\subsection{Experimental setting}\label{subsec:setting}

\textbf{Datasets.}
Three commonly used benchmark datasets are employed to evaluate the effectiveness and robustness of PERO: USTC-TFC \citep{wang2017malware}, ISCX-VPN-Service \citep{gil2016characterization}, and ISCX-VPN-App \citep{gil2016characterization}.
Further details can be found in Appendix \ref{datasets}.

\textbf{Evaluation Metrics.}
We utilize four standard metrics to evaluate the performance of our method: Accuracy (AC), Precision (PR), Recall (RC), and F1-score (F1) \cite{telikani2021cost,zhang2023automatic,azab2024network,chi2024does}.
Accuracy is reported at the sample level, while Precision, Recall, and F1-score are macro-averaged across classes.
To assess robustness, we further evaluate all metrics on the $(1-\alpha)$ tail subset identified by the $\mathrm{CVaR}_{\alpha}$ criterion, yielding tail-conditional scores denoted as $\text{AC}_{\alpha}$, $\text{PR}_{\alpha}$, $\text{RC}_{\alpha}$, and $\text{F1}_{\alpha}$.
Higher values indicate better classification performance.

\textbf{Implementation Details.}
We adopt ET-BERT \cite{lin2022bert} as the backbone and implement the pre-evaluation module as a lightweight MLP for efficient risk estimation.
To validate the generality across backbones, we further evaluate PERO with YaTC \cite{zhao2023yet}.
Detailed architectural configurations and implementation settings are provided in Appendix \ref{appendix:details}.

\subsection{Comparison with Baseline Methods}
\label{subsec:comparison}

\begin{figure}[ht]
\begin{center}
\centerline{\includegraphics[width=0.49\textwidth]{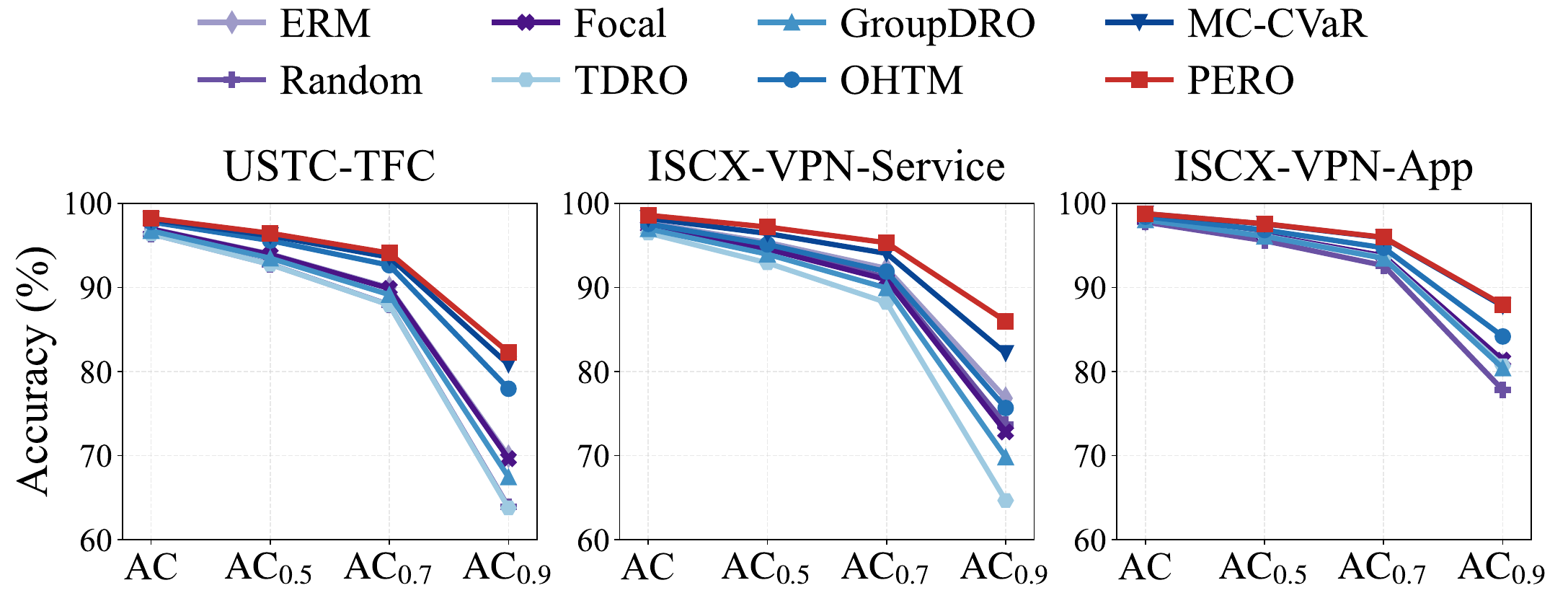}}
\vspace{-0.2cm}
\caption{Comparison of accuracy results across different datasets using various methods.}
\label{fig:accuracy}
\end{center}
\vspace{-13pt}
\end{figure}

We compare PERO with representative baselines spanning standard training, sample selection, loss reweighting, and robust post-training: ERM~\cite{shai2014understanding}, Random Selection, Focal Loss~\cite{lin2017focal}, MC-CVaR~\cite{rockafellar2000optimization}, GroupDRO~\cite{Sagawa2020Distributionally}, OHTM~\cite{kumar2023effect}, and TDRO~\cite{gladin2025improved}.
Baseline details are provided in Appendix~\ref{appendix:baseline}.
The main ET-BERT results on the three benchmarks are shown in Figs.~\ref{fig:accuracy}--\ref{fig:runtime} and Tables~\ref{tab:vpn}--\ref{tab:efficiency}.
Additional YaTC results are reported in Appendix~\ref{appendix:additional_exp}.

\begin{table*}[ht]
\centering
\caption{Comparison Results on ISCX-VPN-Service and ISCX-VPN-App.}
\label{tab:vpn}
\resizebox{\textwidth}{!}{
\begin{tabular}{l | *{4}{c} | *{4}{c} | *{4}{c} | *{4}{c} }
\toprule
Dataset & \multicolumn{8}{c|}{ISCX-VPN-Service} & \multicolumn{8}{c}{ISCX-VPN-App} \\
\midrule
Method & AC  & PR  & RC & F1 & $\text{AC}_{0.9}$& $\text{PR}_{0.9}$& $\text{RC}_{0.9}$ & $\text{F1}_{0.9}$ & AC  & PR  & RC & F1 & $\text{AC}_{0.9}$& $\text{PR}_{0.9}$& $\text{RC}_{0.9}$ & $\text{F1}_{0.9}$\\
\midrule
ERM &97.68 &97.71 &97.68 & 97.69&76.83 & 67.05 &62.13 & 59.61&98.08 & 96.49&96.71 &96.51 &80.80 &72.12 &66.42 &67.93 \\
Random &97.37  &97.39  &97.37  & 97.37 &73.67  & 55.45 &57.45  & 53.21 &97.78  & 96.27 &96.24  &96.12  &77.82  &68.21  &64.12 &65.20  \\
MC-CVaR &\underline{98.22}&\underline{98.24}&\underline{98.22} & \underline{98.22}&\underline{82.17} & \underline{74.33}&\underline{70.37}  & \underline{70.74} &\underline{98.78}  & \textbf{97.54} &\underline{97.77}  &\underline{97.62}  &\underline{87.81}  &\underline{83.27}  &\underline{77.16} &\underline{78.93}  \\
Focal &97.45  &97.48  &97.45  & 97.46 &74.46  & 58.84 &58.25  & 56.39 &98.13  & 96.59 &96.94  &96.72  &81.35  &72.87  &68.37 &69.34  \\
OHTM  &97.57  &97.58  &97.57  & 97.57 &75.67  & 64.66 &66.63  & 64.60 &98.42  & 96.97 &97.44  &97.18  &84.17  &75.68  &70.09 &71.77  \\
GroupDRO &96.98  &97.03  &96.98  & 96.99 &69.83  & 54.59 &54.65  & 52.95 &98.16  & 96.50 &96.84  &96.59  &80.42  &70.47  &64.25 &65.98  \\
TDRO  &96.47  &96.49  &96.47  & 96.47 &64.67  & 55.88 &56.18  & 53.73 &98.25  & 96.79 &96.99  &96.79  &82.51  &73.93  &68.15 &69.81  \\
\midrule
PERO (Ours) &\textbf{98.60}  &\textbf{98.62}  &\textbf{98.60}  & \textbf{98.60} &\textbf{86.00}  & \textbf{83.61} &\textbf{76.42}  & \textbf{78.03} &\textbf{98.79}  & \underline{97.48} &\textbf{97.96}  &\textbf{97.71}  &\textbf{87.94}  &\textbf{85.72}  &\textbf{79.70} &\textbf{81.60}  \\
\bottomrule
\multicolumn{17}{l}{\footnotesize\itshape \textbf{Note:} Results are percentages.
Best and second-best values are in bold and underlined, respectively.} \\
\end{tabular}
}
\end{table*}

\begin{table*}[ht]
\centering
\caption{Comparison Results on USTC-TFC.}
\label{tab:USTC-TFC}

\begin{minipage}{0.6\textwidth}
\centering

\resizebox{\linewidth}{!}{
\begin{tabular}{l | *{4}{c} | *{4}{c}}
\toprule
Dataset & \multicolumn{8}{c}{USTC-TFC} \\
\midrule
Method & AC & PR & RC & F1
& $\text{AC}_{0.9}$ & $\text{PR}_{0.9}$
& $\text{RC}_{0.9}$ & $\text{F1}_{0.9}$ \\
\midrule
ERM
& 96.99 & 97.03 & 97.05 & 97.04
& 69.93 & 33.90 & 32.99 & 31.29 \\

Random
& 96.39 & 96.49 & 96.39 & 96.38
& 63.89 & 45.37 & 41.07 & 41.52 \\

MC-CVaR
& \underline{98.08}
& \underline{98.13}
& \underline{98.13}
& \underline{98.12}
& \underline{80.84}
& \underline{72.01}
& \underline{82.65}
& \underline{75.73} \\

Focal
& 96.96 & 97.04 & 97.02 & 97.03
& 69.65 & 52.71 & 54.40 & 51.82 \\

OHTM
& 97.80 & 97.90 & 97.85 & 97.83
& 77.96 & 71.25 & 76.16 & 72.87 \\

GroupDRO
& 96.75 & 96.80 & 96.83 & 96.81
& 67.46 & 46.87 & 48.55 & 47.13 \\

TDRO
& 96.38 & 96.44 & 96.47 & 96.40
& 63.79 & 49.37 & 52.31 & 48.94 \\
\midrule

PERO (Ours)
& \textbf{98.23}
& \textbf{98.29}
& \textbf{98.27}
& \textbf{98.26}
& \textbf{82.29}
& \textbf{88.38}
& \textbf{87.86}
& \textbf{87.63} \\
\bottomrule
\end{tabular}
}

\vspace{2pt}

\raggedright
\scriptsize\itshape
\textbf{Note:} Results are percentages.
Best and second-best values are in bold and underlined, respectively.

\end{minipage}
\end{table*}

\textbf{Overall and Tail Performance Comparison.}
Fig.~\ref{fig:accuracy} compares the accuracy of different methods across three benchmarks.
Although all methods achieve strong overall accuracy, their performance drops on the $(1-\alpha)$ tail-risk subset, with larger degradation as $\alpha$ increases from 0.7 to 0.9.
PERO and MC-CVaR consistently achieve the strongest overall and tail-conditioned accuracy, clearly outperforming the other baselines.
At $\alpha=0.9$, PERO slightly outperforms MC-CVaR on USTC-TFC and shows a clearer advantage on ISCX-VPN-Service, while the two methods are nearly indistinguishable on ISCX-VPN-App.
Tail degradation is most pronounced on USTC-TFC, where baselines such as ERM, Random, Focal, TDRO, and GroupDRO drop sharply under extreme risk.
By contrast, PERO effectively mitigates tail degradation and achieves the highest $\text{AC}_{0.9}$ across all three datasets.

Tables \ref{tab:vpn}-\ref{tab:USTC-TFC} report overall and $(1-0.9)$-tail-conditioned accuracy, macro precision, macro recall, and macro F1-score.
Consistent with Fig.~\ref{fig:accuracy}, PERO achieves superior or competitive performance across all datasets, especially under tail-conditioned metrics.
On ISCX-VPN-Service, PERO achieves the best $\text{F1}_{0.9}$ and improves over ERM, Random, Focal, OHTM, GroupDRO, TDRO, and MC-CVaR by 18.42\%, 24.82\%, 21.64\%, 13.43\%, 25.08\%, 24.30\%, and 7.29\%, respectively.
On ISCX-VPN-App, PERO also obtains the best tail-conditioned metrics, improving $\text{F1}_{0.9}$ over the same methods by 13.67\%, 16.40\%, 12.26\%, 9.83\%, 15.62\%, 11.79\%, and 2.67\%, respectively, while trailing MC-CVaR by only 0.06\% in overall $\text{PR}$.
The advantages of PERO are most critical on USTC-TFC, where it achieves an $\text{F1}_{0.9}$ of 87.63\%, outperforming the strongest baselines MC-CVaR (75.73\%) and OHTM (72.87\%) and far outperforming Focal (51.82\%), TDRO (48.94\%), GroupDRO (47.13\%), Random (41.52\%), and ERM (31.29\%).
These results show that average-risk training, random selection, focal reweighting, distributionally robust training, and CVaR optimization all struggle on extreme tail samples, whereas PERO maintains strong tail robustness across all three datasets.

Overall, PERO demonstrates superior generalization and tail robustness compared to existing baselines.
By leveraging a lightweight proxy, it matches or exceeds state-of-the-art performance with minimal overhead, suggesting its potential as a low-overhead solution for practical post-training of foundation models.

\textbf{Computational and Memory Efficiency.}
We evaluate the computational and memory efficiency of all methods by comparing their average per-iteration wall-clock runtime and GPU memory usage.
Fig.~\ref{fig:runtime} summarizes the runtime results across three benchmark datasets.
MC-CVaR incurs the highest computational cost, exhibiting the longest runtime on all three datasets.
In contrast, PERO is among the most efficient robustness-oriented methods, consistently running faster than all other robustness-oriented baselines (TDRO, GroupDRO, OHTM, Focal, and MC-CVaR) and remaining close to the lightweight ERM and Random baselines.
This indicates that PERO improves tail-risk robustness without incurring substantial computational overhead.

\begin{figure}[ht]
\begin{center}
\centerline{\includegraphics[width=0.48\textwidth]{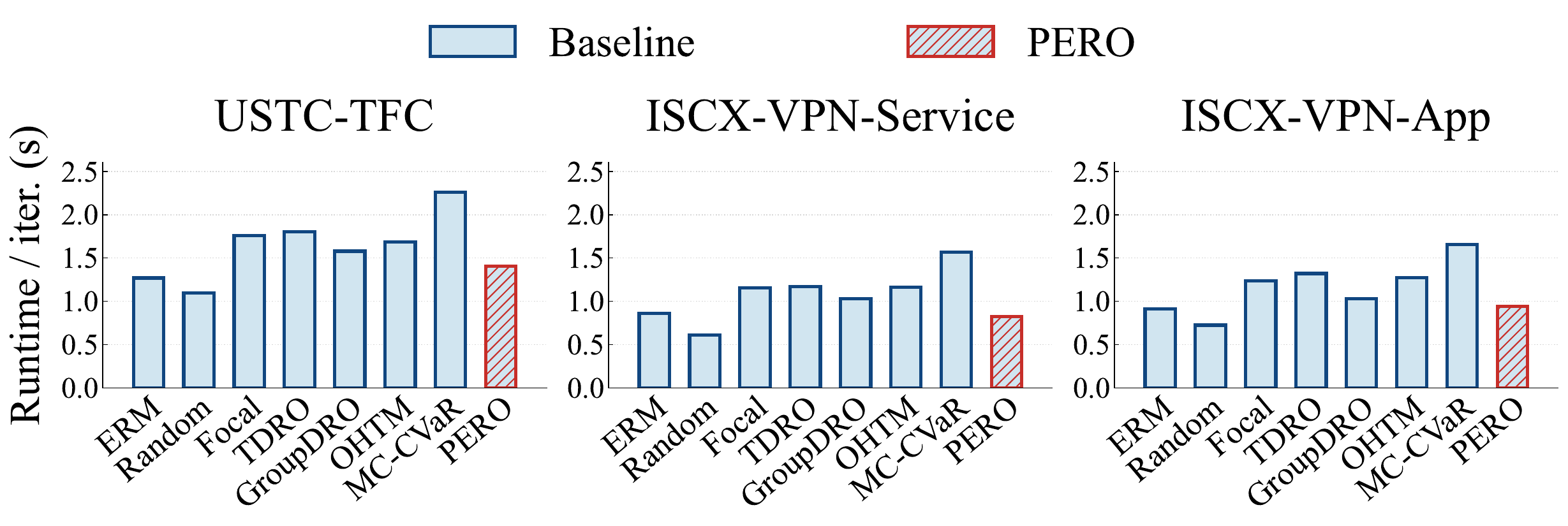}}
\vspace{-0.1cm}
\caption{Runtime comparison across three datasets.
}
\label{fig:runtime}
\end{center}
\vspace{-12pt}
\end{figure}

Table~\ref{tab:efficiency} further reports detailed efficiency statistics on ISCX-VPN-Service.
PERO substantially reduces post-training time, requiring approximately half the runtime of MC-CVaR.
This efficiency gain stems from delegating risk estimation to a lightweight pre-evaluation module and restricting optimization to high-risk samples, thereby avoiding unnecessary forward-backward passes over the full batch.
Effectively, PERO shifts the bulk of computation from large model backpropagation to small model inference, addressing the real bottleneck in foundation model post-training.
In terms of memory usage, PERO consumes 11.75 GB of GPU memory, compared to 26.18 GB for MC-CVaR, achieving a 55.1\% reduction.
While ERM, Focal Loss, and TDRO incur the lowest memory footprint, PERO remains comparable, with only a 0.16 GB difference.

\begin{table}[t]
\centering
\caption{Efficiency Comparison Results on ISCX-VPN-Service.}
\label{tab:efficiency}
\scalebox{0.9}{
\begin{tabular}{lcc}
\toprule
Method & Runtime / Iter. (s) & Memory Usage (GB) \\
\midrule
ERM & 0.864 & 11.59 \\
Random & 0.610 & 11.82 \\
MC-CVaR & 1.568 & 26.18 \\
Focal & 1.156 & 11.59 \\
OHTM & 1.164 & 17.27 \\
GroupDRO & 1.032 & 11.63 \\
TDRO & 1.170 & 11.59 \\
\midrule
\textbf{PERO (Ours)} & \textbf{0.826} & \textbf{11.75} \\
\bottomrule
\end{tabular}}
\vspace{-3pt}
\end{table}

\subsection{Risk Prediction Fidelity of the Pre-evaluation Module}
\label{subsec:predict}

We assess the pre-evaluation module as a surrogate risk estimator using Pearson correlation, Spearman correlation, and Precision@$16$, where $16=B/2$ under batch size $B=32$.
These capture, respectively, linear agreement between predicted and exact losses, the consistency of their risk rankings, and the overlap between predicted and exact top-$16$ high-risk samples.

\begin{figure}[ht]
\begin{center}
\centerline{\includegraphics[width=0.48\textwidth]{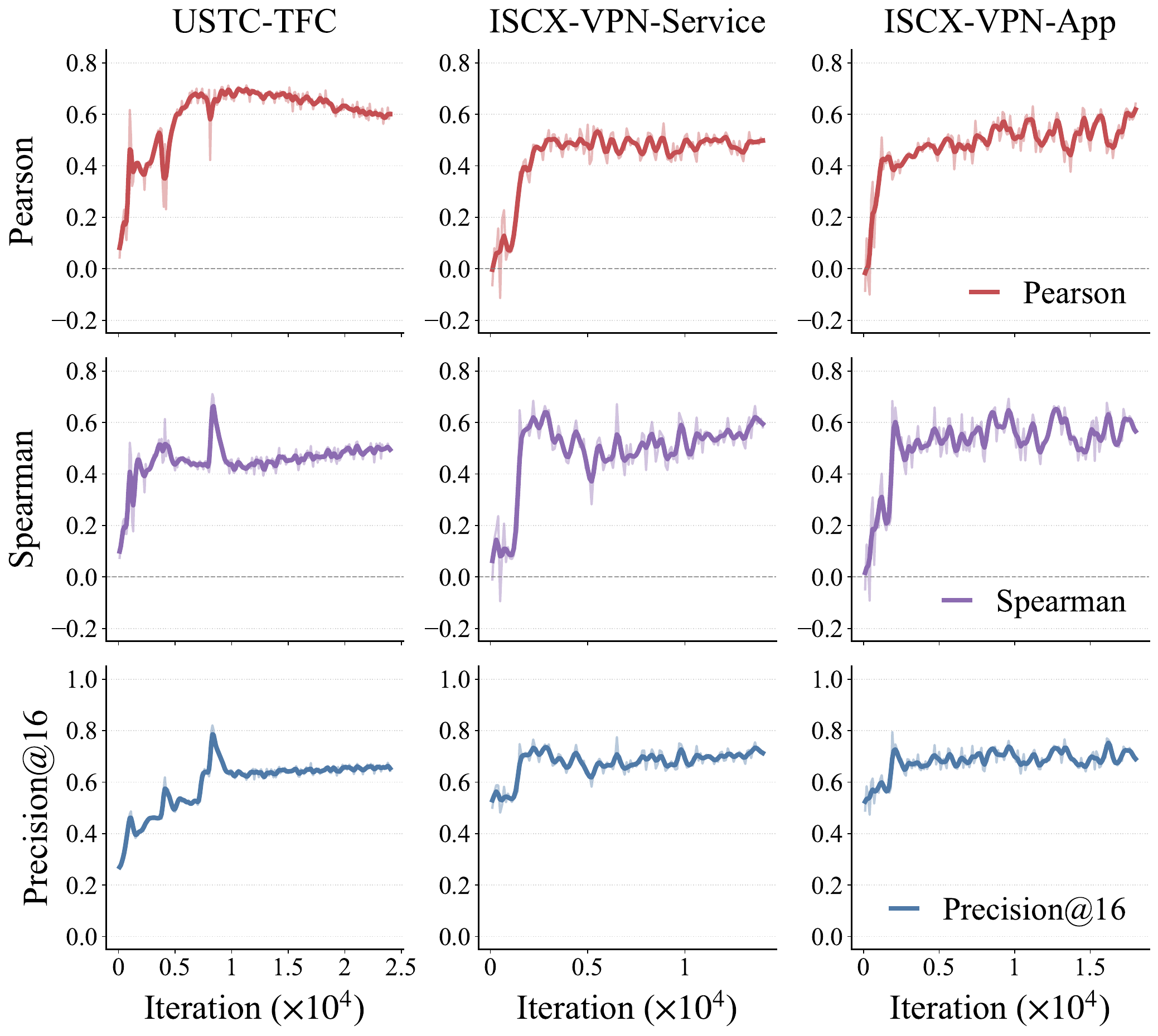}}
\vspace{-0.1cm}
\caption{Pearson correlation, Spearman correlation, and Precision@16 between predicted and exact risks during training across three benchmark datasets.}
\label{fig:correlation}
\end{center}
\vspace{-12pt}
\end{figure}

As shown in Fig.~\ref{fig:correlation}, Pearson and Spearman correlations rise from weak early-stage values and stabilize at moderate levels, typically around $0.5$ to $0.6$ with dataset-dependent variation.
Precision@$k$ also shows that many top-ranked samples selected by the module overlap with the true highest-loss samples, stabilizing around $0.65$ to $0.7$ in the later training stage.
The initially low values are expected, as rapid early classifier updates weaken the alignment between exact losses and proxy predictions.
Their stabilization indicates that the module captures informative relative risk patterns.
Together, these results show that the pre-evaluation module provides sufficient functional fidelity, in both ranking and top-set identification, to support effective robust post-training.

\begin{figure}[t]
\begin{center}
\centerline{\includegraphics[width=0.48\textwidth]{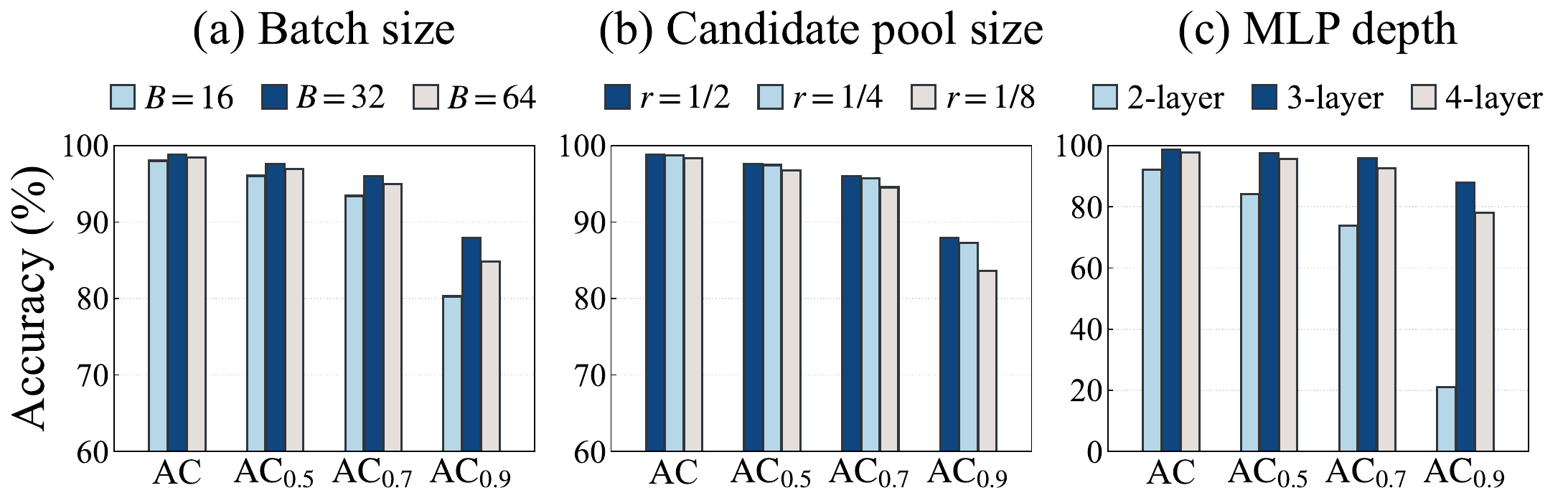}}
\vspace{-0.1cm}
\caption{Ablation study of key PERO hyperparameters.
(a) Training batch size $B$.
(b) Candidate pool size $\hat{B}$ and the selection ratio $r = B / \hat{B}$.
(c) MLP depth in the pre-evaluation module.
Dark blue bars indicate PERO's default settings.
}
\label{fig:ablation}
\end{center}
\vspace{-12pt}
\end{figure}

\subsection{Ablation Study}\label{subsec:ablation}
We evaluate PERO's sensitivity to key hyperparameters: the training batch size $B$, the candidate pool size $\hat{B}$, and the depth of the MLP within the prediction head in the pre-evaluation module.
Each hyperparameter is varied independently from the default settings of $B=32$, $\hat{B}=64$, and a 3-layer MLP.
All experiments are conducted on ISCX-VPN-App, with results summarized in Fig.~\ref{fig:ablation}.

\textbf{Effect of batch size $B$.}
We vary $B \in \{16,32,64\}$.
Fig.~\ref{fig:ablation}(a) shows that increasing $B$ from 16 to 32 improves all metrics, particularly tail accuracy $\text{AC}_{0.9}$, while further increasing $B$ to 64 yields diminishing returns.
These results suggest that a moderate batch size provides a better trade-off between optimization stability and tail sensitivity, while overly small or large batches are less effective.
Notably, PERO remains strong under non-optimal batch sizes, outperforming all baselines except MC-CVaR at $B=64$ and staying competitive with ERM and GroupDRO at $B=16$.

\textbf{Effect of candidate pool size $\hat{B}$.}
We vary $\hat{B} \in \{64,128,256\}$, using $r = B/\hat{B}$ to denote the selection ratio.
The case $\hat{B}=B$ is omitted as it degenerates to ERM.
As shown in Fig.~\ref{fig:ablation}(b), all sizes achieve near-perfect performance at $\text{AC}$, $\text{AC}_{0.5}$, and $\text{AC}_{0.7}$.
At higher risk levels, larger selection ratios consistently retain a slight advantage, with the gap becoming more noticeable at $\text{AC}_{0.9}$.
This suggests that sufficient exploration benefits both overall performance and tail robustness, whereas excessively small selection ratios mildly degrade robustness.
In particular, at $r=1/4$, PERO achieves the best $\text{AC}_{0.9}$ among all baselines except MC-CVaR, while maintaining comparable performance to MC-CVaR.
At $r=1/8$, it retains its lead over ERM and GroupDRO and matches OHTM.

\textbf{Effect of MLP depth.}
We evaluate MLP depths of 2, 3, and 4 layers for the prediction head in the pre-evaluation module.
As shown in Fig.~\ref{fig:ablation}(c), the 3-layer MLP achieves the best performance across all risk levels.
The 2-layer variant provides a lighter surrogate but has more limited expressive capacity, especially under the most stringent tail-risk setting.
Increasing the depth to 4 layers does not bring further gains, suggesting that additional surrogate capacity is unnecessary for this task.
Overall, the 3-layer MLP offers the best trade-off between risk estimation capacity, robustness, and efficiency, and the 4-layer variant remains competitive with ERM and GroupDRO on $\text{AC}_{0.9}$.
This underscores that moderate model complexity suffices for strong tail performance without requiring aggressive surrogate capacity.

\section{Conclusion}

This work proposes PERO, an efficient robust post-training framework for encrypted traffic foundation models.
PERO introduces a lightweight pre-evaluation module that decouples sample-wise risk estimation from expensive foundation-model optimization, enabling targeted updates on critical tail examples with low computational overhead.
Experiments on three main benchmarks show that PERO consistently improves tail robustness over representative robust post-training baselines while substantially reducing computational and memory costs.
Additional results with YaTC further support the scalability and backbone generality of PERO.
These efficiency gains are particularly important for continual post-training scenarios, where limited update windows make direct CVaR-style optimization impractical.
Beyond encrypted traffic classification, PERO may extend to other risk-sensitive large-scale learning settings with high update costs and limited budgets, which warrants further validation in future work.



\appendix
\section*{Appendix}

\section{Limitations and Ethical Considerations}\label{appendix:limit}
\textbf{Data Privacy and Consent.}
We use publicly released encrypted traffic datasets  provided by the original authors for research purposes.
PERO does not inspect communication content, attempt payload decryption, or reconstruct user identities or behaviors.

\textbf{Fairness and Bias.}
PERO improves tail-risk robustness during post-training by emphasizing high-risk samples, but it does not remove biases inherent in the training data.
Tail-risk performance may therefore vary across network environments.

\textbf{Potential Misuse.}
Encrypted traffic classification technology may be misused for intrusive monitoring if deployed without adequate security measures.
Responsible use requires transparency and appropriate access safeguards.

\textbf{Limitations.}
Our theoretical convergence result (Theorem \ref{thm:convergence}) relies on a uniformly bounded loss-Hessian spectral radius throughout optimization.
Although this is a standard smoothness condition in convergence analyses of gradient-based methods \cite{hu2023beyond}, it is strong and may not strictly hold for large Transformer models in general.
Thus, Theorem \ref{thm:convergence} should be interpreted as analytical insight under explicit simplifying conditions rather than a complete characterization of practical optimization dynamics.
Although the supplementary CICIoT2022 experiments evaluate PERO on a larger and more realistic dataset, all results are still based on publicly available benchmarks, which may not fully capture temporal drift and deployment dynamics in production encrypted traffic environments.
The pre-evaluation module also provides approximate risk estimates rather than exact losses, which may affect subset selection under severe distribution shifts.
Finally, PERO introduces additional hyperparameters, motivating future work on adaptive parameter selection and validation in more diverse network environments.

\section{Dataset Details}\label{datasets}

We evaluate PERO on multiple encrypted traffic classification benchmarks.
The main ET-BERT experiments use USTC-TFC, ISCX-VPN-Service, and ISCX-VPN-App.
For backbone validation, we further apply PERO to YaTC on USTC-TFC, ISCX-VPN, ISCX-Tor, and CICIoT2022.

\begin{itemize}
    \item USTC-TFC~\cite{wang2017malware} contains 97,115 encrypted traffic samples from 10 malware families and 10 benign applications.

    \item ISCX-VPN~\cite{gil2016characterization} contains encrypted traffic under VPN and non-VPN scenarios.
    Following ET-BERT, we use ISCX-VPN-Service with 60,000 samples and 12 service-level classes, and ISCX-VPN-App with 77,163 samples and 17 application-level classes.
    For YaTC-based experiments, we follow the YaTC setting and report results on ISCX-VPN.

    \item ISCX-Tor~\cite{lashkari2017characterization} contains Tor traffic across eight categories.

    \item CICIoT2022~\cite{dadkhah2022towards} covers diverse IoT device types and attack behaviors, and is used as a large-scale IoT traffic benchmark.
\end{itemize}

\begin{table*}[!ht]
\centering
\caption{Supplementary Results on ISCX-VPN and ISCX-Tor with YaTC as Backbone.}
\label{addtional_tab:vpn}

\resizebox{\textwidth}{!}{
\begin{tabular}{l | *{4}{c} | *{4}{c} | *{4}{c} | *{4}{c} }
\toprule
Dataset & \multicolumn{8}{c|}{ISCX-VPN} & \multicolumn{8}{c}{ISCX-Tor} \\
\midrule
Method & AC  & PR  & RC & F1 & $\text{AC}_{0.9}$& $\text{PR}_{0.9}$& $\text{RC}_{0.9}$ & $\text{F1}_{0.9}$ & AC  & PR  & RC & F1 & $\text{AC}_{0.9}$& $\text{PR}_{0.9}$& $\text{RC}_{0.9}$ & $\text{F1}_{0.9}$\\
\midrule
ERM  
&97.37 &96.80 &95.50 &95.97 &74.14 &\underline{77.49} &71.54 &72.23 
&98.62 &97.71 &98.81 &98.18 &\underline{86.49} &87.61 &\underline{91.45} &86.50 \\

Random 
&96.50 &96.24 &94.27 &95.15 &65.52 &59.65 &54.16 &54.79 
&98.08 &96.88 &98.33 &97.46 &81.08 &87.79 &87.38 &84.81 \\

MC-CVaR 
&97.54 &\underline{97.28} &95.56 &96.35 &75.86 &70.67 &\textbf{78.13} &71.37 
&\textbf{98.90} &\textbf{98.36} &\textbf{99.16} &\textbf{98.71} &\textbf{89.19} &87.50 &91.27 &\underline{87.01} \\

Focal 
&97.20 &95.83 &95.61 &95.61 &72.41 &62.97 &58.32 &59.62 
&98.35 &97.25 &98.67 &97.86 &83.78 &\textbf{91.63} &83.33 &84.56 \\

OHTM 
&97.20 &96.46 &95.79 &96.00 &72.41 &53.94 &50.71 &51.17 
&98.35 &97.46 &98.67 &97.98 &83.78 &84.13 &88.47 &83.44 \\

GroupDRO 
&97.55 &96.51 &95.20 &95.70 &75.86 &76.57 &75.38 &\underline{74.25} 
&\underline{98.63} &97.84 &98.51 &98.14 &\underline{86.49} &76.56 &80.21 &76.84 \\

TDRO 
&\underline{97.72} &97.18 &\underline{96.05} &\underline{96.44} &\underline{77.59} &69.26 &65.71 &64.85 
&98.35 &97.42 &98.29 &97.80 &83.78 &75.00 &76.38 &73.61 \\
\midrule
PERO (Ours) 
&\textbf{97.90} &\textbf{97.68} &\textbf{96.32} &\textbf{96.88} &\textbf{79.31} &\textbf{82.22} &\underline{78.05} &\textbf{76.56} 
&\underline{98.63} &\underline{97.86} &\underline{98.90} &\underline{98.32} &\underline{86.49} &\underline{87.93} &\textbf{91.67} &\textbf{87.87} \\
\bottomrule
\multicolumn{17}{l}{\footnotesize\itshape \textbf{Note:} Results are percentages.
Best and second-best values are in bold and underlined, respectively.} \\
\end{tabular}
}
\end{table*}

\begin{table*}[!ht]
\centering
\caption{Supplementary Results on USTC-TFC and CICIoT2022 with YaTC as Backbone.}
\label{addtional_tab:USTC-TFC}

\resizebox{\textwidth}{!}{ 
\begin{tabular}{l | *{4}{c} | *{4}{c} | *{4}{c} | *{4}{c} }
\toprule
Dataset & \multicolumn{8}{c|}{USTC-TFC}  & \multicolumn{8}{c}{CICIoT2022} \\
\midrule
Method & AC  & PR  & RC & F1 & $\text{AC}_{0.9}$ & $\text{PR}_{0.9}$ & $\text{RC}_{0.9}$ & $\text{F1}_{0.9}$ 
& AC  & PR  & RC & F1 & $\text{AC}_{0.9}$& $\text{PR}_{0.9}$& $\text{RC}_{0.9}$ & $\text{F1}_{0.9}$\\
\midrule
ERM  
&97.05 &97.87 &97.74 &97.80 &70.71 &84.03 &85.57 &84.68 
&95.55 &94.85 &\underline{93.26} &\underline{93.97} &55.56 &65.16 &59.44 &60.32 \\

Random 
&96.94 &97.79 &97.67 &97.72 &69.70 &78.01 &81.83 &79.28 
&95.32 &94.83 &92.79 &93.66 &53.28 &64.07 &62.85 &61.84 \\

MC-CVaR 
&\underline{97.35} &98.09 &97.96 &98.02 &73.73 &87.10 &\textbf{88.07} &\textbf{87.54} 
&95.61 &\textbf{95.69} &92.32 &93.63 &56.13 &\textbf{73.17} &\underline{65.78} &\textbf{67.13} \\

Focal 
&96.54 &97.57 &96.44 &96.91 &65.66 &\underline{87.40} &84.61 &84.76 
&95.23 &94.67 &91.89 &92.91 &52.13 &62.42 &62.05 &59.78 \\

OHTM 
&96.84 &97.72 &97.59 &97.65 &68.69 &84.03 &84.89 &84.43 
&95.27 &\underline{95.50} &92.56 &93.74 &52.71 &62.05 &58.01 &58.61 \\

GroupDRO 
&\underline{97.35} &\underline{98.11} &\underline{97.98} &\underline{98.03} &\underline{73.74} &\textbf{89.94} &\underline{87.47} &86.97 
&\underline{95.78} &95.24 &92.64 &93.67 &\underline{57.83} &61.86 &59.26 &59.52 \\

TDRO  
&97.25 &98.01 &97.89 &97.95 &72.72 &86.48 &87.22 &86.83 
&95.67 &94.44 &91.93 &93.03 &56.70 &62.56 &54.43 &55.87 \\
\midrule
PERO (Ours) 
&\textbf{97.56} &\textbf{98.24} &\textbf{98.11} &\textbf{98.17} &\textbf{75.76} &86.87 &87.33 &\underline{87.09}  
&\textbf{96.01} &94.90 &\textbf{94.02} &\textbf{94.39} &\textbf{60.11} &\underline{66.50} &\textbf{67.69} &\underline{64.84} \\
\bottomrule
\multicolumn{17}{l}{\footnotesize\itshape \textbf{Note:} Results are percentages.
Best and second-best values are in bold and underlined, respectively.} \\
\end{tabular}
}
\end{table*}

\section{Implementation Details}\label{appendix:details}
\textbf{Model Architecture.}
We use pre-trained ET-BERT~\cite{lin2022bert} as the primary backbone because it is widely adopted, achieves strong cross-benchmark performance, and has a large parameter scale that highlights the efficiency benefit of lightweight risk estimation.
We further validate backbone generality with YaTC~\cite{zhao2023yet} in supplementary experiments.

The pre-evaluation module is a lightweight regression network.
It encodes sequence embeddings with a two-layer MLP followed by global mean pooling, embeds hard labels via a lookup table, and projects soft probability targets through a linear layer.
The concatenated representations are then passed to a three-layer MLP head with dimensions 770, 385, and 385 to regress the loss value.

\textbf{Data Preprocessing and Splits.}
For ET-BERT experiments, we use three publicly available datasets from prior ET-BERT studies,\footnote{\url{https://github.com/linwhitehat/ET-BERT}} following the same preprocessing and tokenization procedures.
For YaTC experiments, we likewise adopt the original preprocessing pipeline.\footnote{\url{https://github.com/NSSL-SJTU/YaTC}}
All datasets are randomly split into training, validation, and test sets with an 8:1:1 ratio, preserving class distributions.

\textbf{Training Protocol.}
The default training batch size is \(B=32\).
For selection-based methods, each batch is expanded to a candidate pool of \(\hat{B}=2B\), unless otherwise specified.
The pre-evaluation module is optimized with Adam using a learning rate of \(5\times10^{-6}\), supervised by historical risks from the primary classifier.
The primary classifier is optimized with AdamW using a learning rate of \(1\times10^{-6}\), following the standard ET-BERT fine-tuning protocol.
Non-selection methods are trained for 10 epochs, while selection-based methods are trained for 20 epochs to keep the number of classifier-updated samples comparable.
All results are averaged over five runs with different random seeds.

\textbf{Baseline Implementations.}\label{appendix:baseline}
All baselines use the same backbone, optimizer, learning rate, training schedule, and batch size as PERO.
Selection-based baselines also use the same candidate pool size \(\hat{B}=64\).

\begin{itemize}
    \item \textbf{Random Selection} uniformly samples \(B\) candidates, serving as a na\"ive selection baseline without risk estimation.

    \item \textbf{MC-CVaR} \cite{rockafellar2000optimization} uses a Monte Carlo approximation with risk level \(\alpha=0.5\), selecting the top-\(B\) highest-loss candidates to approximate conditional tail-risk optimization.

    \item \textbf{Focal Loss}~\cite{lin2017focal} re-weights each loss by $(1{-}p_t)^\gamma$ to emphasize hard examples, without subset selection.

    \item \textbf{TDRO}~\cite{gladin2025improved} minimizes a LogSumExp surrogate that smoothly up-weights high-loss samples, without subset selection.

    \item \textbf{GroupDRO} \cite{Sagawa2020Distributionally} is adapted to the instance level by assigning softmax weights over detached per-sample losses and optimizing the weighted loss sum, without group labels or a candidate pool.

    \item \textbf{OHTM} \cite{kumar2023effect} selects candidates by greedy Gram--Schmidt orthogonalization over \(\ell_2\)-normalized embeddings, encouraging geometrically diverse subsets.

    \item \textbf{ERM} \cite{shai2014understanding} trains on each mini-batch with standard empirical risk minimization, without re-weighting or selection.
\end{itemize}

\textbf{Hardware and Code Availability.}
All experiments run on NVIDIA A100 GPUs with PyTorch.
The complete implementation will be released upon acceptance.
Core code is available at \url{https://anonymous.4open.science/r/PERO-D0A7}.

\textbf{Potential Extension to RL Post-training}
Although PERO is developed for supervised post-training~\cite{zou2025utility,zou2026flylora}, its pre-evaluation and selective optimization paradigm may naturally extend to LLMs~\cite{chi2024unveiling,lv2026breaking} and RL post-training~\cite{guo2025deepseek, qu2026can, mao2026dynamics,zou2026trace}. Specifically, a lightweight proxy could predict the optimization utility of sampled trajectories, allowing the policy to focus expensive updates on the most informative trajectories.

\section{Additional Backbone Evaluation}\label{appendix:additional_exp}

To further assess scalability and backbone generality, we conduct additional experiments using YaTC \cite{zhao2023yet} as the backbone.
Following the evaluation setting in YaTC, we evaluate PERO on four benchmarks: USTC-TFC \cite{wang2017malware}, ISCX-VPN \cite{gil2016characterization}, ISCX-Tor \cite{lashkari2017characterization}, and CICIoT2022 \cite{dadkhah2022towards}, with results reported in Tables~\ref{addtional_tab:vpn}--\ref{addtional_tab:USTC-TFC}.

PERO obtains the best F1 on ISCX-VPN, USTC-TFC, and CICIoT2022, and remains within 0.39\% of the best F1 on ISCX-Tor.
Under the tail-conditioned metrics, PERO achieves the best $\text{AC}_{0.9}$ on three of the four datasets and the best $\text{F1}_{0.9}$ on ISCX-VPN and ISCX-Tor.
In particular, on CICIoT2022, PERO improves $\text{AC}_{0.9}$ from the second-best 57.83\% to 60.11\%.
Although MC-CVaR obtains the highest $\text{F1}_{0.9}$ due to stronger tail precision, PERO achieves the best $\text{AC}_{0.9}$ and $\text{RC}_{0.9}$, suggesting that PERO provides broader tail coverage.
Overall, while MC-CVaR and GroupDRO occasionally lead on individual metrics, PERO provides a more balanced trade-off between classification performance and tail robustness, confirming its effectiveness under a different backbone.

\newpage

\bibliography{main_bib_reduced}
\bibliographystyle{icml2021}

\end{document}